\documentclass[10pt,twocolumn,letterpaper]{article}

\usepackage{cvpr}              % To produce the CAMERA-READY version
\usepackage{graphicx}
\usepackage{amsmath}
\usepackage{amssymb}
\usepackage{booktabs}
\usepackage{rotating}
\usepackage{multirow}
\usepackage{hhline}
\usepackage{diagbox}
\usepackage{makecell}

\usepackage{pdfrender}
\usepackage{amsbsy}

\usepackage[pagebackref,breaklinks,colorlinks]{hyperref}
\usepackage[capitalize]{cleveref}
\crefname{section}{Sec.}{Secs.}
\Crefname{section}{Section}{Sections}
\Crefname{table}{Table}{Tables}
\crefname{table}{Tab.}{Tabs.}

\def\cvprPaperID{8244} % *** Enter the CVPR Paper ID here
\def\confName{CVPR}
\def\confYear{2023}

\begin{document}

\title{ViTs for SITS: Vision Transformers for Satellite Image Time Series}

\author{{Michail Tarasiou \qquad Erik Chavez \qquad Stefanos Zafeiriou} \\ Imperial College London \\
{\tt\small \{michail.tarasiou10,  erik.chavez, s.zafeiriou}\}{\tt\small @imperial.ac.uk}}

\maketitle

%%%%%%%%% ABSTRACT
\begin{abstract}
  In this paper we introduce the Temporo-Spatial Vision Transformer (TSViT), a fully-attentional model for general Satellite Image Time Series (SITS) processing based on the Vision Transformer (ViT). TSViT splits a SITS record into non-overlapping patches in space and time which are tokenized and subsequently processed by a factorized temporo-spatial encoder. We argue, that in contrast to natural images, a temporal-then-spatial factorization is more intuitive for SITS processing and present experimental evidence for this claim. Additionally, we enhance the model's discriminative power by introducing two novel mechanisms for acquisition-time-specific temporal positional encodings and multiple learnable class tokens. The effect of all novel design choices is evaluated through an extensive ablation study. Our proposed architecture achieves state-of-the-art performance, surpassing previous approaches by a significant margin in three publicly available SITS semantic segmentation and classification datasets. All model, training and evaluation codes can be found at {\color{magenta} \small \verb{https://github.com/michaeltrs/DeepSatModels{\color{black}.}
\end{abstract}

\vspace{-0.3cm} 

\section{Introduction} \label{sec:intro}

\begin{figure}[!t]
     % \centering
     \hspace{0mm}
     \begin{subfigure}[b]{0.45\textwidth}
         \centering
   \includegraphics[width=0.8\linewidth]{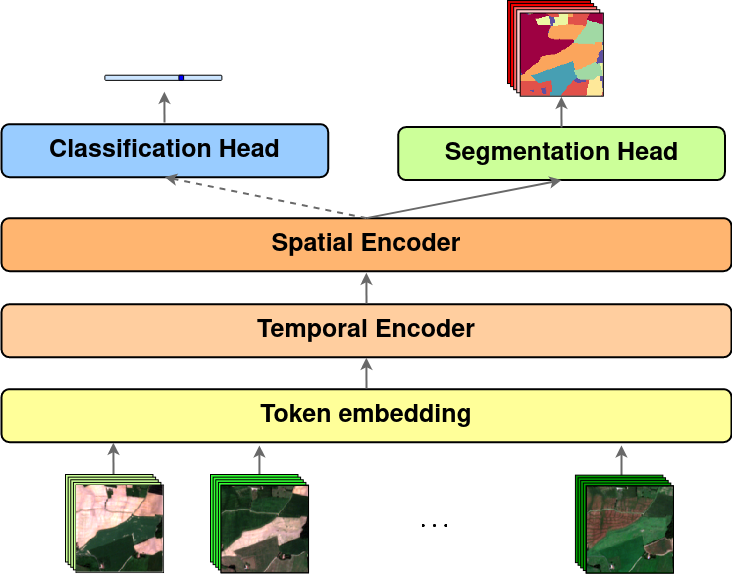}
        %  \caption{}
         % \label{fig:model_overview}
     \end{subfigure}
     \hfill
     \vspace{5mm}
     \begin{subfigure}[b]{0.45\textwidth}
         \centering
   \includegraphics[width=0.9\linewidth]{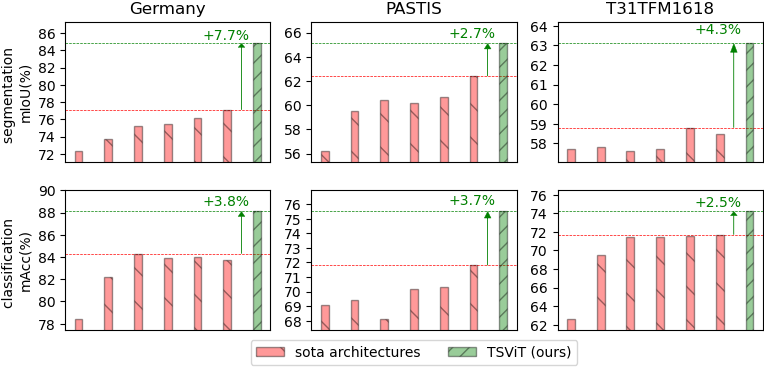}
        %  \caption{$y=3sinx$}
         % \label{fig:three sin x}
     \end{subfigure}
     \hfill
     \caption{{\bf Model and performance overview. (top)} TSViT architecture. A more detailed schematic is presented in Fig.\ref{fig:TSViT_submodules}. {\bf (bottom) TSViT performance} compared with previous arts (Table \ref{tab:results}).}
     \label{fig:overview}
\end{figure}

The monitoring of the Earth surface man-made impacts or activities is essential to enable the design of effective interventions to increase welfare and resilience of societies. One example is the sector of agriculture in which monitoring of crop development can help design optimum strategies aimed at improving the welfare of farmers and resilience of the food production system. 
The second of United Nations Sustainable Development Goals (SDG) of Ending Hunger relies on increasing the crop productivity and revenues of farmers in poor and developing countries \cite{ungoals} - approximately 2.5 billion people's livelihoods depend mainly on producing crops \cite{Conway2012}. Achieving SDG 2 goals requires to be able to accurately monitor yields and the evolution of cultivated areas in order to measure the progress towards achieving several goals, as well as to evaluate the effectiveness of different policies or interventions. 
In the European Union (EU) the Sentinel for Common Agricultural Policy program (Sen4CAP) \cite{sen4cap} focuses on developing tools and analytics to support the verification of direct payments to farmers with underlying environmental conditionalities such as the adoption of environmentally-friendly \cite{practices} and crop diversification \cite{diversification} practices based on real-time monitoring by the European Space Agency's (ESA) Sentinel high-resolution satellite constellation \cite{EUa} to complement on site verification. 
Recently, the volume and diversity of space-borne Earth Observation (EO) data \cite{50yearsEO} and post-processing tools \cite{googleearth,deepsatdata,TrainingDML-AI} has increased exponentially. This wealth of resources, in combination with important developments in machine learning for computer vision \cite{alexnet, resnet, fcn}, provides an important opportunity for the development of tools for the automated monitoring of crop development. 

Towards more accurate automatic crop type recognition, we introduce TSViT, the first fully-attentional\footnote{without any convolution operations} architecture for general SITS processing. An overview of the proposed architecture can be seen in Fig.\ref{fig:overview} (top). Our novel design introduces some inductive biases that make TSViT particularly suitable for the target domain:
\begin{itemize}
    \item Satellite imagery for monitoring land surface variability boast a high revisit time leading to long temporal sequences. To reduce the amount of computation we factorize input dimensions into their temporal and spatial components, providing intuition (section \ref{sec:tsvit_encoder}) and experimental evidence (section \ref{sec:ablation}) about why the order of factorization matters. 
    \item TSViT uses a Transformer backbone \cite{aiayn} following the recently proposed ViT framework \cite{vit}. As a result, every TSViT layer has a global receptive field in time or space, in contrast to previously proposed convolutional and recurrent architectures \cite{Ruwurm2,Rustowicz2019SemanticSO,garnot_iccv,duplo,tempcnn}.
    \item To make our approach more suitable for SITS modelling we propose a tokenization scheme for the input image timeseries and propose acquisition-time-specific temporal position encodings in order to extract date-aware features and to account for irregularities in SITS acquisition times (section \ref{sec:position_encodings}). 
    \item We make modifications to the ViT framework (section \ref{sec:tsvit_backbone}) to enhance its capacity to gather class-specific evidence which we argue suits the problem at hand and design two custom decoder heads to accommodate both global and dense predictions (section \ref{sec:tsvit_decoder}).
\end{itemize}
Our provided intuitions are tested through extensive ablation studies on design parameters presented in section \ref{sec:ablation}. 
Overall, our architecture achieves state-of-the-art performance in three publicly available datasets for classification and semantic segmentation presented in Table \ref{tab:results} and Fig.\ref{fig:overview}.

%------------------------------------------------------------------------
\section{Related work}
\label{sec:related_work}

\subsection{Crop type recognition}

Crop type recognition is a subcategory of land use recognition which involves assigning one of $K$ crop categories (classes) at a set of desired locations on a geospatial grid. For successfully doing so modelling the temporal patterns of growth during a time period of interest has been shown to be critical \cite{temp_var,spacetime}. As a result, model inputs are timeseries of  $T$ satellite images of spatial dimensions $H \times W$ with $C$ channels, $\mathbf{X} \in \mathbb{R}^{T \times H\times W \times C}$ rather than single acquisitions. 
There has been a significant body of work on crop type identification found in the remote sensing literature \cite{cc1, ndvi1, ndvi2, ndvi3, hmm, pel_rand}. These works typically involve multiple processing steps and domain expertise to guide the extraction of features, e.g. NDVI \cite{dvi}, that can be separated into crop types by learnt classifiers. More recently, Deep Neural Networks (DNN) trained on raw optical data \cite{Ruwurm1,Ru_wurm_2018,dnn2,dnn3,dnn4,emb_earth} have been shown to outperform these approaches.
At the object level, (SITS classification) \cite{tempcnn, duplo, transformer_sat} use 1D data of single-pixel or parcel-level aggregated feature timeseries, rather than the full SITS record, learning a mapping $f: \mathbb{R}^{T \times C} \rightarrow \mathbb{R}^{K}$. Among these works, TempCNN \cite{tempcnn} employs a simple 1D convolutional architecture, while \cite{transformer_sat} use the Transformer architecture \cite{aiayn}. DuPLo \cite{duplo} consists of an ensemble of CNN and RNN streams in an effort to exploit the complementarity of extracted features. Finally, \cite{garnot2019satellite} view satellite images as un-ordered sets of pixels and calculate feature statistics at the parcel level, but, in contrast to previously mentioned approaches, their implementation requires knowledge of the object geometry.
At the pixel level (SITS semantic segmentation), models learn a mapping $f(\mathbf{X}) \in \mathbb{R}^{H \times W \times K}$. For this task, \cite{Ru_wurm_2018} show that convolutional RNN variants (CLSTM, CGRU) \cite{conv_lstm} can automatically extract useful features from raw optical data, including cloudy images, that can be linearly separated into classes. The use of CNN architectures is explored in \cite{Rustowicz2019SemanticSO} who employ two models: a UNET2D feature extractor, followed by a CLSTM temporal model (UNET2D-CLSTM); and a UNET3D fully-convolutional model. Both are found to achieve equivalent performances. In a similar spirit, \cite{fpn_clstm} use a FPN \cite{fpn} feature extractor, coupled with a CLSTM temporal model (FPN-CLSTM). The UNET3Df architecture \cite{cscl} follows from UNET3D but uses a different decoder head more suited to contrastive learning. The U-TAE architecture \cite{garnot_iccv} follows a different approach, in that it employs the encoder part of a UNET2D, applied on parallel on all images, and a subsequent temporal attention mechanism which collapses the temporal feature dimension. These spatial-only features are further processed by the decoder part of a UNET2D to obtain dense predictions.

\subsection{Self-attention in vision}

Convolutional \cite{alexnet,vgg,resnet} and fully-convolutional networks (FCN) \cite{overfeat,fcn} have been the de-facto model of choice for vision tasks over the past decade. The convolution operation extracts translation-equivariant features via application of a small square kernel over the spatial extent of the learnt representation and grows the feature receptive field linearly over the depth of the network. In contrast, the self-attention operation, introduced as the main building block of the Transformer architecture \cite{aiayn}, uses self-similarity as a means for feature aggregation and can have a global receptive field at every layer. Following the adoption of Transformers as the dominant architecture in natural language processing tasks \cite{aiayn,bert,gpt3}, several works have attempted to exploit self-attention in vision architectures. Because the time complexity of self-attention scales quadratically with the size of the input, its naive implementation on image data, which typically contain more pixels than text segments contain words, would be prohibitive. To bypass this issue, early works focused on improving efficiency by injecting self-attention layers only at specific locations within a CNN \cite{nonlocal,attn_augm_cnn} or by constraining their receptive field to a local region \cite{image_transformer,stand_sa,axial_deeplab}, however, in practice, these designs do not scale well with available hardware leading to slow throughput rates, large memory requirements and long training times. 
Following a different approach, the Vision Transformer (ViT) \cite{vit}, presented in further detail in section \ref{sec:vit_backbone}, constitutes an effort to apply a pure Transformer architecture on image data, by proposing a simple, yet efficient image tokenization strategy. 
Several works have drawn inspiration from ViT to develop novel attention-based architectures for vision. For image recognition, \cite{tokens2token,swin_transformer} re-introduce some of the inductive biases that made CNNs successful in vision, leading to improved performances without the need for long pre-training schedules, \cite{dense_vit,segmenter} employ Transformers for dense prediction, \cite{detr, vit_yolo, song2022vidt} for object detection and \cite{video_transformer,video_instance_vit,vivit} for video processing. 
Among these works, our framework is more closely related to \cite{vivit} who use ViT for video processing. However, we deviate significantly from their design by introducing a dictionary of acquisition-time-specific position encodings to accommodate an uneven distribution of images in time, by employing a different input factorization strategy more suitable to SITS, and by being interested in both global and dense predictions (section \ref{sec:tsvit_encoder}). Finally, we are using a multi-token strategy that allows better handling of spatial interactions and leads to improved class separation. Multiple {\it cls} tokens have also been employed in \cite{detr,segmenter}. However, while both studies use them as class queries inputs to a decoder module, we introduce the {\it cls} tokens as an input to the encoder in order to collapse the time dimension and obtain class-specific features. 

\section{Method}
In this section we present the TSViT architecture in detail. First, we give a brief overview of the ViT (section \ref{sec:vit_backbone}) which provided inspiration for this work. In section \ref{sec:tsvit_backbone} we present our modified TSViT backbone, followed by our tokenization scheme (section \ref{sec:tokenization}), encoder (section \ref{sec:tsvit_encoder}) and decoder (section \ref{sec:tsvit_decoder}) modules. Finally, in section \ref{sec:position_encodings}, we discuss several considerations behind the design of our position encoding scheme. 

\begin{figure}
   \centering
   \includegraphics[width=0.9\linewidth]{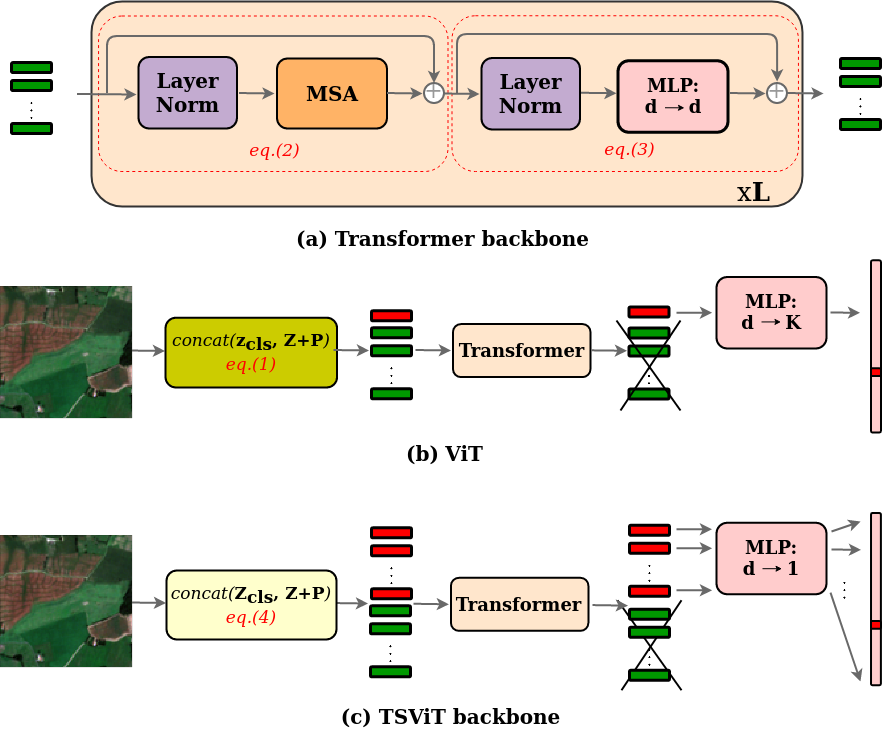}
   \caption{{\bf Backbone architectures. (a)} Transformer backbone, {\bf (b)} ViT architecture, {\bf (c)} TSViT backbone employs additional {\it cls} tokens (red), each responsible for predicting a single class. }
    \label{fig:vit}
\end{figure}

\subsection{Primer on ViT} \label{sec:vit_backbone}
Inspired by the success of Transformers in natural language processing tasks \cite{aiayn} the ViT \cite{vit} is an application of the Transformer architecture to images with the fewest possible modifications. Their framework involves the tokenization of a 2D image $\mathbf{X} \in \mathbb{R}^{H\times W \times C}$ to a set of {\it patch} tokens $\mathbf{Z} \in \mathbb{R}^{N \times d}$ by splitting it into a sequence of $N=\lfloor\frac{H}{h}\rfloor \lfloor\frac{W}{w}\rfloor$ same-size and non-overlapping patches of spatial extent $(h \times w )$ which are flattened into 1D tokens $\mathbf{x_i} \in \mathbb{R}^{hwC}$ and linearly projected into $d$ dimensions. Overall, the process of token extraction is equivalent to the application of 2D convolution with kernel size $(h \times w )$ at strides $(h, w)$ across respective dimensions. The extracted patches are used to construct model inputs as follows:

\begin{equation}\label{eq:vit_inputs}
    \mathbf{Z^0} = concat(\mathbf{z_{cls}}, \mathbf{Z} + \mathbf{P}) \in \mathbb{R}^{N+1\times d}
\end{equation}

\noindent A set of learned positional encoding vectors $\mathbf{P} \in \mathbb{R}^{N\times d}$, added to $\mathbf{Z}$, are employed to encode the absolute position information of each token and break the permutation invariance property of the subsequent Transformer layers. A separate learned class ({\it cls}) token $\mathbf{z_{cls}} \in \mathbb{R}^{d}$ \cite{bert} is prepended to the linearly transformed and positionally augmented {\it patch} tokens leading to a length $N+1$ sequence of tokens $\mathbf{Z^0}$ which are used as model inputs. The Transformer backbone consists of $L$ blocks of alternating layers of Multiheaded Self-Attention (MSA) \cite{aiayn} and residual Multi-Layer Perceptron (MLP) (Fig.\ref{fig:vit}(a)). 

\begin{equation}\label{block1}
    \mathbf{Y^l} = MSA(LN(\mathbf{Z^l})) + \mathbf{Z^l}
\end{equation}

\begin{equation}\label{block2}
    \mathbf{Z^{l+1}} = MLP(LN(\mathbf{Y^l})) + \mathbf{Y^l}
\end{equation}

\noindent Prior to each layer, inputs are normalized following Layernorm (LN) \cite{ln}. MLP blocks consist of two layers of linear projection followed by GELU non-linear activations \cite{gelu}. In contrast to CNN architectures, in which spatial dimensions are reduced while feature dimensions increase with layer depth, Transformers are isotropic in that all feature maps  $\mathbf{Z}^l \in \mathbb{R}^{1+N\times d}$ have the same shape throughout the network. % The final feature map $\mathbf{Z^L}$ 
After processing by the final layer L, all tokens apart from the first one (the state of the {\it cls} token) are discarded and unormalized class probabilities are calculated by processing this token via a MLP. A schematic representation of the ViT architecture can be seen in Fig.\ref{fig:vit}(b).

\begin{figure}[!t]
   \centering

   \includegraphics[width=0.8\linewidth]{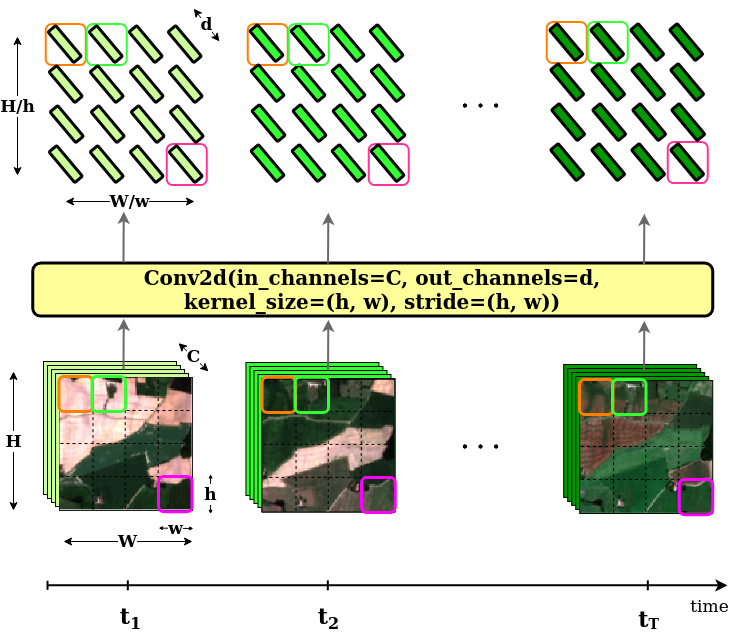}
   \caption{{\bf SITS Tokenization}. We embed each satellite image independently following ViT \cite{vit}}
    \label{fig:tokenization}
\end{figure}

\subsection{Backbone architecture}\label{sec:tsvit_backbone}
In the ViT architecture, the {\it cls} token progressively refines information gathered from all {\it patch} tokens to reach a final global representation used to derive class probabilities. 

Our TSViT backbone, shown in Fig.\ref{fig:vit}(c), essentially follows from ViT, with modifications in the tokenization and decoder layers. More specifically, we introduce $K$ (equal to the number of object classes) additional learnable {\it cls} tokens $\mathbf{Z_{cls}} \in \mathbb{R}^{K \times d}$, compared to ViT which uses a single token. 
\begin{equation}\label{eq:tsvit_inputs}
    \mathbf{Z^0} = concat(\mathbf{Z_{cls}}, \mathbf{Z} + \mathbf{P}) \in \mathbb{R}^{N+K\times d}
\end{equation}

Without deviating from ViT, all {\it cls} and positionally augmented {\it patch} tokens are concatenated and processed by the $L$ layers of a Transformer encoder. After the final layer, we discard all {\it patch} tokens and project each {\it cls} token into a scalar value. By concatenating these values we obtain a length $K$ vector of unormalised class probabilities. 
This design choice brings the following two benefits: 1) it increases the capacity of the {\it cls} token relative to the {\it patch} tokens, allowing them to store more patterns to be used by the MSA operation; introducing multiple {\it cls} tokens can be seen as equivalent to increasing the dimension of a single {\it cls} token to an integer multiple of the {\it patch} token dimension $d_{cls} = k d_{patch}$ and split the {\it cls} token into $k$ separate subspaces prior to the MSA operation. In this way we can increase the capacity of the {\it cls} tokens while avoiding issues such as the need for asymmetric MSA weight matrices for {\it cls} and {\it patch} tokens, which would effectively more than double our model's parameter count.
2) it allows for more controlled handling of the spatial interactions between classes. By choosing $k=K$ and enforcing a bijective mapping from {\it cls} tokens to class predictions, the state of each {\it cls} token becomes more focused to a specific class with network depth. In TSViT we go a step further and explicitly separate {\it cls} tokens by class after processing with the temporal encoder to allow only same-{\it cls}-token interactions in the spatial encoder. In section \ref{sec:tsvit_encoder} we argue why this is a very useful inductive bias for modelling spatial relationships in crop type recognition. 

\subsection{Tokenization of SITS inputs}\label{sec:tokenization}
A SITS record $\mathbf{X} \in \mathbb{R}^{T \times H\times W \times C}$ consists of a series of $T$ satellite images of spatial dimensions $H \times W$ with $C$ channels. 
For the tokenization of our 3D inputs, we can extend the tokenization-as-convolution approach to 3D data and apply a 3D kernel with size $(t \times h \times w)$ at stride $(t,h,w)$ across temporal and spatial dimensions. In this manner $N=\lfloor\frac{T}{t}\rfloor \lfloor\frac{H}{h}\rfloor \lfloor\frac{W}{w}\rfloor$ non-overlapping tokens $\mathbf{x_i} \in \mathbb{R}^{thwC}$ are extracted, and subsequently projected to $d$ dimensions. 
% introduce N_H, N_W, N_T. where $N_T=\lfloor\frac{T}{t}\rfloor, N_H= \lfloor\frac{H}{h}\rfloor, N_W= \lfloor\frac{W}{w}\rfloor$
Using $t>1$, all extracted tokens contain  spatio-temporal information. For the special case of $t=1$ each token contains spatial-only information for each acquisition time and temporal information is accounted for only through the encoder layers. Since the computation cost of global self-attention layers is quadratic w.r.t. the length of the token sequence $\mathcal{O}(N^2)$, choosing larger values for $t,h,w$ can lead to significantly reduced number of FLOPS. In our experiments, however, we have found small values for $t,h,w$ to work much better in practice. For all presented experiments we use a value of $t=1$ motivated in part because this choice simplifies the implementation of acquisition-time-specific temporal position encodings, described in section \ref{sec:position_encodings}. With regards to the spatial dimensions of extracted patches we have found small values to work best for semantic segmentation, which is reasonable given that small patches retain additional spatial granularity. In the end, our tokenization scheme is similar to ViT's applied in parallel for each acquisition as shown in Fig.\ref{fig:tokenization}, however, at this stage, instead of unrolling feature dimensions, we retain the spatial structure of the original input as reshape operations will be handled by the TSViT encoder submodules.
% 3x3 spatial x 1 temporal. Cost.
% {\color{red} COMMENT ON COST INCREASE}
% \subsection{Vision transformers for SITS processing}
% \subsection{Temporal-Spatial Vision Transformer (TSViT)}

\begin{figure*}
   \centering
   \includegraphics[width=0.975\linewidth]{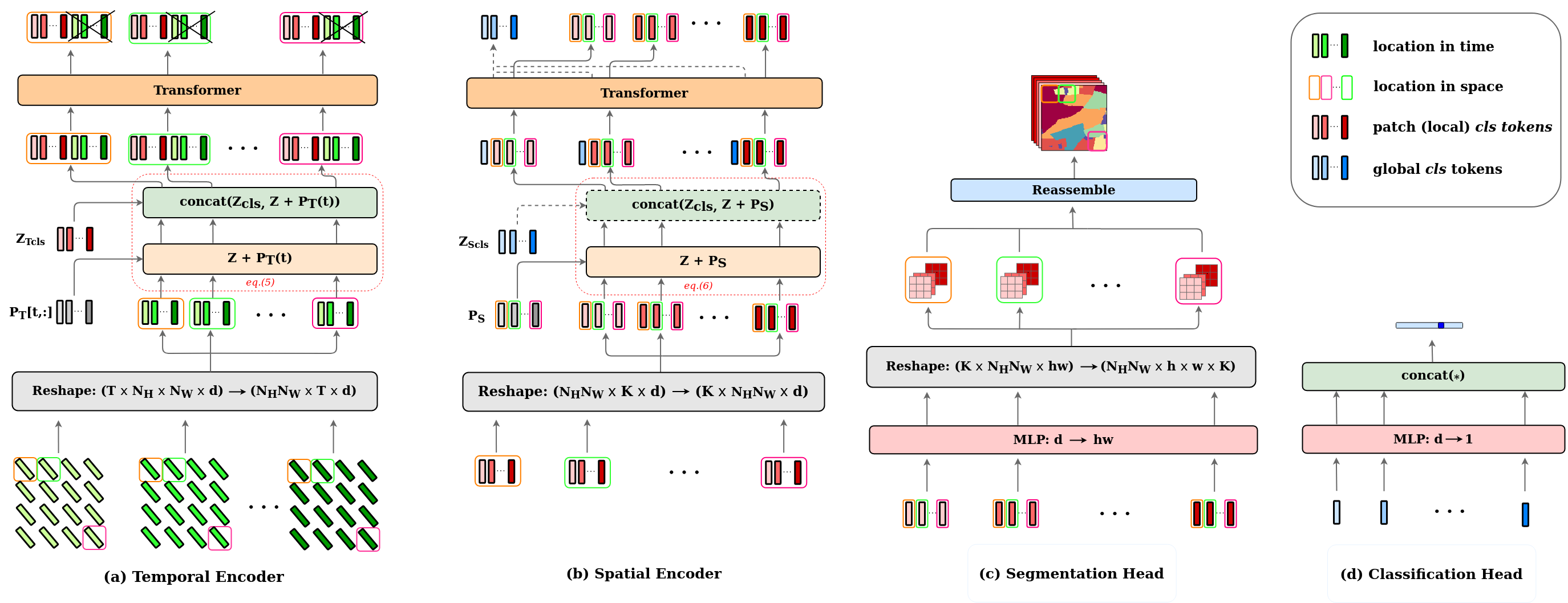}
   \caption{{\bf TSViT submodules. (a)} Temporal encoder. We reshape tokenized inputs, retaining the spatio-temporal structure of SITS, into a set of timeseries for each spatial location, add temporal position encodings $\mathbf{P_T[t,:]}$ for acquisition times $\mathbf{t}$, concatenate local {\it cls} tokens $\mathbf{Z_{Tcls}}$ (eq.\ref{eq:temp_transf_input}) and process in parallel with a Transformer. Only the first $K$ output tokens are retained. {\bf (b)} Spatial encoder. We reshape the outputs of the temporal encoder into a set of spatial feature maps for each {\it cls} token, add spatial position encodings  $\mathbf{P_S}$, concatenate global {\it cls} tokens $\mathbf{Z_{Scls}}$ (eq.\ref{eq:spatial_encoder_input}) and process in parallel with a Transformer. {\bf (c)} Segmentation head. Each local {\it cls} token is projected into $hw$ values denoting class-specific evidence for every pixel in a patch. All patches are then reassembled into the original image dimensions. {\bf (d)} Classification head. Global {\it cls} tokens are projected into scalar values, each denoting evidence for the presence of a specific class.}
   \label{fig:TSViT_submodules}
\end{figure*}

\subsection{Encoder architecture}\label{sec:tsvit_encoder}
In the previous section we presented a motivation for using small values $t,h,w$ for the extracted patches. Unless other measures are taken to reduce the model's computational cost this choice would be prohibitive for processing SITS with multiple acquisition times. To avoid such problems, we choose to factorize our inputs across their temporal and spatial dimensions, a practice commonly employed for video processing \cite{spatial_temp1,spatial_temp2,spatial_temp3,spatial_temp4,spatial_temp5,spatial_temp6}. We note that all these works use a spatial-temporal factorization order, which is reasonable when dealing with natural images, given that it allows the extraction of higher level, semantically aware spatial features, whose relationship in time is useful for scene understanding. However, we argue that in the context of SITS, reversing the order of factorization is a meaningful design choice for the following reasons: 
1) in contrast to natural images in which context can be useful for recognising an object, in crop type recognition context can provide little information, or can be misleading. This arises from the fact that the shape of agricultural parcels, does not need to follow its intended use, i.e. most crops can generally be cultivated independent of a field's size or shape. Of course there exist variations in the shapes and sizes of agricultural fields \cite{agri_land_patterns}, but these depend mostly on local agricultural practices and are not expected to generalize over unseen regions. Furthermore, agricultural parcels do not inherently contain sub-components or structure. Thus, knowing what is cultivated in a piece of land is not expected to provide information about what grows nearby. This is in contrast to other objects which clearly contain structure, e.g. in human face parsing there are clear expectations about the relative positions of various face parts. We test this hypothesis in the supplementary material by enumerating over all agricultural parcels belonging to the most popular crop types in the T31TFM {\it S2} tile in France and taking crop-type-conditional pixel counts over a 1km square region from their centers. Then, we calculate the cosine similarity of these values with unconditional pixel counts over the extent of the T31TFM tile and find a high degree of similarity, suggesting that there are no significant variations between these distributions; 
2) a small region in SITS is far more informative than its equivalent in natural images, as it contains more channels than regular RGB images ({\it S2} imagery contains 13 bands in total) whose intensities are averaged over a relatively large area (highest resolution of {\it S2} images is $10 \times 10$ m$^2$); 
3) SITS for land cover recognition do not typically contain moving objects. As a result, a timeseries of single pixel values can be used for extracting features that are informative of a specific object part found at that particular location. 
Therefore, several objects can be recognised using only information found in a single location; plants, for example, can be recognised by variations of their spectral signatures during their growth cycle. Many works performing crop classification do so using only temporal information in the form of timeseries of small patches \cite{Ru_wurm_2018}, pixel statistics over the extent of parcels \cite{Ruwurm1} or even values from single pixels \cite{transformer_sat, tempcnn}. 
% Several works have quantified the \cite{spacetime}
On the other hand, the spatial patterns in a single image are uninformative of the crop type, as evidenced by the low performance of systems relying on single images \cite{garnot_iccv}. Our encoder architecture can be seen in Fig.\ref{fig:TSViT_submodules}(a,b). We now describe the temporal and spatial encoder submodules.
% Object locations are fixed in the image. As a result the same pixel will contain information 
 
% but is motivated by facts such as land availability and its optimal division into  
% , as the choice of how agricultural land is used is independent of location  
% The temporal encoder processes interactions across all acquisitions for the same patch.

{\bf Temporal encoder} Thus, we tokenize a SITS record $\mathbf{X} \in \mathbb{R}^{T \times H\times W \times C}$ into a set of tokens of size $(N_T \times N_H \times N_W  \times d)$, as described in section \ref{sec:tokenization} and subsequently reshape to $\mathbf{Z_T} \in \mathbb{R}^{N_H N_W \times N_T  \times d}$, to get a list of token timeseries for all patch locations. The input to the temporal encoder is:
%$\mathbf{Z_T^0} \in \mathbb{R}^{N_H N_W \times K+N_{T}  \times d}$ with:
% dense vs cls TSViT
% \begin{equation}\label{eq:temp_transf_input}
%     \mathbf{{Z^0_T}_i} = concat(\mathbf{z_{cls}}, [\mathbf{z_{i1}}, \mathbf{z_{i2}}, ..., \mathbf{z_{iN_T}}] + \mathbf{P_T(t)}) \in \mathbb{R}^{N_T+1\times d}
% \end{equation}
% Where $\mathbf{z_{cls}} \in \mathbb{R}^d$ is a {\it cls} token and $\mathbf{P_T(t)} \in \mathbb{R}^{N_T  \times d}$ represents learned position encodings.
% % By processing $\mathbf{Z^0}$ via the temporal transformer we obtain: 
% The input to the temporal transformer becomes:
\begin{equation}\label{eq:temp_transf_input}
    \mathbf{{Z^0_T}} = concat(\mathbf{Z_{Tcls}}, \mathbf{Z_T} + \mathbf{P_T[t,:]}) \in \mathbb{R}^{N_HN_W \times K+N_T\times d}
\end{equation}
% [\mathbf{z_{i1}}, \mathbf{z_{i2}}, ..., \mathbf{z_{iN_T}}]
where $\mathbf{P_T[t,:]} \in \mathbb{R}^{N_T  \times d}$ and $\mathbf{Z_{Tcls}} \in \mathbb{R}^{K \times d}$ are respectively added and prepended to all $N_HN_W$ timeseries and $\mathbf{t} \in \mathbb{R}^T$ is a vector containing all $T$ acquisition times. 
All samples are then processed in parallel by a Transformer module. Consequently, the final feature map of the temporal encoder becomes $\mathbf{Z^L_T} \in \mathbb{R}^{N_H N_W \times K + N_T  \times d}$ in which the first $K$ tokens in the temporal dimension correspond to the prepended {\it cls} tokens. We only keep these tokens, discarding the remaining $N_T$ vectors.

{\bf Spatial encoder} We now transpose the first and second dimensions in the temporal encoder output, to obtain a list of patch features $\mathbf{Z_S} \in \mathbb{R}^{K \times N_H N_W \times d}$ for all output classes. In a similar spirit, the input to the spatial encoder becomes:

\begin{equation}\label{eq:spatial_encoder_input}
    \mathbf{{Z^0_S}} = concat(\mathbf{Z_{Scls}}, \mathbf{Z_S} + \mathbf{P_S}) \in \mathbb{R}^{K \times 1 + N_H N_W \times d}
\end{equation}
where $\mathbf{P_S} \in \mathbb{R}^{N_H N_W \times d}$ are respectively added to all $K$ spatial representations and each element of $\mathbf{Z_{Scls}} \in \mathbb{R}^{K \times 1 \times d}$ is prepended to each class-specific feature map. We note, that while in the temporal encoder {\it cls} tokens were prepended to all patch locations, now there is a single {\it cls} token per spatial feature map such that $\mathbf{Z_{Scls}}$ are used to gather global SITS-level information. Processing with the spatial encoder leads to a similar size output feature map $\mathbf{{Z^L_S}} \in \mathbb{R}^{K \times 1 + N_H N_W \times d}$. 

\subsection{Decoder architecture}\label{sec:tsvit_decoder}
The TSViT encoder architecture described in the previous section is designed as a general backbone for SITS processing. To accommodate both global and dense prediction tasks we design two decoder heads which feed on different components of the encoder output. We view the output of the encoder as $\mathbf{{Z^L_S}} = [\mathbf{{Z^L_{Sglobal}}} | \mathbf{{Z^L_{Slocal}}}]$ respectively corresponding to the states of the global and local {\it cls} tokens.
For {\bf image classification}, we only make use of $\mathbf{{Z^L_{Sglobal}}} \in \mathbb{R}^{K \times d}$. We proceed, as described in sec.\ref{sec:tsvit_backbone}, by projecting each feature into a scalar value and concatenate these values to obtain global unormalised class probabilities as shown in Fig.\ref{fig:TSViT_submodules}(d).
Complementarily, for {\bf semantic segmentation} we only use $\mathbf{{Z^L_{Slocal}}} \in \mathbb{R}^{K \times N_HN_W \times d}$. These features encode information for the presence of each class over the spatial extent of each image patch. By projecting each feature into $hw$ dimensions and further reshaping the feature dimension to $(h \times w)$ we obtain a set of class-specific probabilities for each pixel in a patch. It is possible now to merge these patches together into an output map $(H \times W \times K)$ which represents class probabilities for each pixel in the original image. This process is presented schematically in Fig.\ref{fig:TSViT_submodules}(c). 
\vspace{-0.1cm}

% \subsection{Positional encodings}\label{position_encodings}
\subsection{Position encodings}\label{sec:position_encodings}
As described in section \ref{sec:tsvit_encoder}, positional encodings are injected in two different locations in our proposed network. First, temporal position encodings are added to all {\it patch} tokens before processing by the temporal encoder (eq.\ref{eq:temp_transf_input}). This operation aims at breaking the permutation invariance property of MSA by introducing time-specific position biases to all extracted {\it patch} tokens. For crop recognition encoding the absolute temporal position of features is important as it helps identifying a plant's growth stage within the crop cycle. Furthermore, the time interval between successive images in SITS varies depending on acquisition times and other factors, such as the degree of cloudiness or corrupted data. To introduce acquisition-time-specific biases into the model, our temporal position encodings $\mathbf{P_T[t,:]}$ depend directly on acquisition times $\mathbf{t}$. More specifically, we make note of all acquisition times $\mathbf{t'} = [t_1, t_2, ..., t_{T'}]$ found in the training data and construct a lookup table $\mathbf{P_T} \in \mathbb{R}^{T' \times d}$ containing all learnt position encodings indexed by date. Finding the date-specific encodings that need to be added to {\it patch} tokens (eq.\ref{eq:temp_transf_input}) reduces to looking up appropriate indices from $\mathbf{P_T}$. In this way temporal position encodings introduce a dynamic prior of where to look at in the models' global temporal receptive field, rather than simply encoding the order of SITS acquisitions which would discard valuable information. 
Following token processing by the temporal encoder, spatial position embeddings $\mathbf{P_S}$ are added to the extracted {\it cls} tokens. These are not dynamic in nature and are similar to the position encodings used in the original ViT architecture, with the difference that these biases are now added to $K$ feature maps instead of a single one.  

\vspace{-0.1cm}
\section{Experiments}\label{sec:experiments}
We apply TSViT to two tasks using SITS records $\mathbf{X} \in \mathbb{R}^{T \times H\times W \times C}$ as inputs: classification and semantic segmentation.  
% For both tasks, model inputs are timeseries of images $\mathbf{X} \in \mathbb{R}^{T \times H\times W \times C}$ (SITS). 
At the object level, classification models learn a mapping $f(\mathbf{X}) \in \mathbb{R}^{K}$ for the object occupying the center of the $H \times W$ region. Semantic segmentation models learn a mapping $f(\mathbf{X}) \in \mathbb{R}^{H \times W \times K}$, predicting class probabilities for each pixel over the spatial extent of the SITS record. We use an ablation study on semantic segmentation to guide model design and hyperparameter tuning and proceed with presenting our main results on three publicly available SITS semantic segmentation and classification datasets. 

\vspace{-0.1cm}

\subsection{Training and evaluation}
{\bf Datasets} To evaluate the performance of our proposed semantic segmentation model we are using three publicly available {\it S2} land cover recognition datasets. 
The dataset presented in \cite{Ru_wurm_2018} covers a densely cultivated area of interest of $102 \times 42$ km$^2$ north of Munich, Germany and contains 17 distinct classes. Individual image samples cover a $240 \times 240$ m$^2$ area ($24 \times 24$ pixels) and contain 13 bands. 
The PASTIS dataset \cite{garnot_iccv} contains images from four different regions in France with diverse climate and crop distributions, spanning over 4000 km$^2$ and including 18 crop types. In total, it includes 2.4k SITS samples of size $128 \times 128$, each containing 33-61 acquisitions and 10 image bands. Because the PASTIS sample size is too large for efficiently training TSViT with available hardware, we split each sample into $24 \times 24$ patches and retain all acquisition times for a total of 60k samples. To accommodate a large set of experiments we only use fold-1 among the five folds provided in PASTIS.
Finally, we use the T31TFM-1618 dataset \cite{cscl} which covers a densely cultivated {\it S2} tile in France for years 2016-18 and includes 20 distinct classes. In total, it includes 140k samples of size $48 \times 48$, each containing 14-33 acquisitions and 13 image bands.
For the SITS classification experiments, we construct the datasets from the respective segmentation datasets. More specifically, for PASTIS we use the provided object instance ids to extract $24 \times 24$ pixel regions whose center pixel falls inside each object and use the class of this object as the sample class. The remaining two datasets contain samples of smaller spatial extent, making the above strategy not feasible in practice. Here, we choose to retain the samples as they are and assign the class of the center pixel as the global class. We note that this strategy forces us to discard samples in which the center pixels belongs to the background class. Additional details are provided in the supplementary material.

{\bf Implementation details} 
For all experiments presented we train for the same number of epochs using the provided data splits from the respective publications for a fair comparison. More specifically, we train on all datasets using the provided training sets and report results on the validation sets for Germany and T31TFM-1618, and on the test set for PASTIS. For training TSViT we use the AdamW optimizer \cite{adamw} with a learning rate schedule which includes a warmup period starting from zero to a maximum value $10^{-3}$ at epoch 10, followed by cosine learning rate decay \cite{loshchilov2017sgdr} down to $5*10^{-6}$ at the end of training. For Germany and T31TFM-1618 we train with the above settings and report the best performances between what we achieve and the original studies. Since we split PASTIS, we are training with both settings and report the best results. Overall, we find that our settings improve model performance. We train with a batch size of 16 or 32 and no regularization on $\times 2$ Nvidia Titan Xp gpus in a data parallel fashion. All models are trained with a Masked Cross-Entropy loss, masking the effect of the background class in both training loss and evaluation metrics. We report overall accuracy (OA), averaged over pixels, and mean intersection over union (mIoU) averaged over classes.
For SITS classification, in addition to the 1D models presented in section \ref{sec:related_work} we modify the best performing semantic segmentation models by aggregating extracted features across space prior to the application of a classifier, thus, outputing a single prediction. Classification models are trained with Focal Loss \cite{focal_loss}. We report OA and mean accuracy (mAcc) averaged over classes.

\begin{table}[!t]
\begin{center}
\begin{tabular}{c|cc|c}
\hline
{\bf Ablation} & \multicolumn{2}{c|}{{\bf Settings}} & {\bf mIoU}  \\
\hline \hline
\multirow{2}{*}{Factorization order}& \multicolumn{2}{c|}{Spatial \& Temporal} & 48.8\\
                                    &\multicolumn{2}{c|}{{\bf Temporal \& Spatial}} & {\bf 78.5}\\%   &   &   &   
\hline
\multirow{2}{*}{\#{\it cls} tokens} & \multicolumn{2}{c|}{1} & 78.5\\
                                    & \multicolumn{2}{c|}{{\bf K}} & {\bf 83.6}\\
\hline
\multirow{2}{*}{Position encodings} & \multicolumn{2}{c|}{Static} & 80.8\\
& \multicolumn{2}{c|}{{\bf Date lookup}}  & {\bf 83.6}\\
\hline
\multirow{4}{*}{ \makecell{Interactions between \\ {\it cls} tokens  }} & Temporal & Spatial & \\
\cline{2-3}
    % &  \tikzxmark &  \tikzxmark & 84.7\\
    % &  \tikzxmark &  \checkmark  & \\
    &  \checkmark &  \checkmark  & 81.5\\
    &  \pmb{\checkmark} &   \pmb{X} & {\bf 83.6}\\
\hline
\multirow{3}{*}{Patch size}& \multicolumn{2}{c|}{$\mathbf{2 \times 2}$} & {\bf 84.8}\\
                           & \multicolumn{2}{c|}{$3 \times 3$} & 83.6\\
                           % & \multicolumn{2}{c|}{$4 \times 4$} & 81.5\\
                           & \multicolumn{2}{c|}{$6 \times 6$} & 79.6\\%   &   &   &   \\
\hline
\end{tabular}
\end{center}
\caption{{\bf Ablation on design choices for TSViT}. All proposed design choices are found to have a positive effect on performance.}
\label{tab:ablation}
\end{table}

% \begin{figure}[!t]
% \begin{center}
% % \fbox{\rule{0pt}{1in} \rule{0.9\linewidth}{0pt}}
%    \includegraphics[width=0.925\linewidth, height=2cm]{figures/static_vs_lookup.png}
% \end{center}
%    \caption{{\bf Date lookup vs Static position encodings} for a varying number of timestamps in Germany. Here T is used as a proxy for .}
% \label{fig:pos_encodings}
% \end{figure}

\begin{table*}[!h]
% \fontsize{6}{8}  %\selectfont text in size 8 pt
% \tiny
\centering
\resizebox{0.95\textwidth}{!}{
\begin{tabular}{lccccc}
\cline{4-6}
 & & & \textbf{Germany} \cite{Ru_wurm_2018} & \textbf{PASTIS} \cite{garnot_iccv} & \textbf{T31TFM-1618} \cite{cscl} \\
 \hline
 \textbf{Model}  & \textbf{$\#$params. (M)}& \textbf{IT (ms)}&    & \textbf{Semantic segmentation (OA / mIoU)} &    \\
 \hline
 BiCGRU \cite{Ru_wurm_2018} & 4.5 & 38.6 &  91.3 / 72.3 & 80.5 / 56.2 & 88.6 / 57.7 \\
 FPN-CLSTM \cite{fpn_clstm}  & 1.2 & 19.5 &  91.8 / 73.7 & 81.9 / 59.5 &  88.4 / 57.8 \\ 
 UNET3D \cite{Rustowicz2019SemanticSO} & 6.2 & 11.2 &  92.4 / 75.2 & 82.3 / 60.4 & 88.4 / 57.6 \\
 UNET3Df \cite{cscl} & 7.2 & 19.7 &  92.4 / 75.4 & 82.1 / 60.2 & 88.6 / 57.7  \\
 UNET2D-CLSTM \cite{Rustowicz2019SemanticSO} & 2.3 & 35.5 &  92.9 / 76.2 & 82.7 / 60.7  & 89.0 / 58.8  \\
 U-TAE \cite{garnot_iccv} & 1.1 & 8.8 &  93.1 / 77.1   & 82.9 / 62.4 (83.2 / 63.1) & 88.9 / 58.5  \\ 
{\bf TSViT (ours)} & 1.7 & 11.8 &  {\bf 95.0 / 84.8} & {\bf 83.4 / 65.1 (83.4 / 65.4)} & {\bf 90.3 / 63.1} \\ 
\hline
 \textbf{Model} & \textbf{$\#$params. (M)} & \textbf{IT (ms)} &  & \textbf{Object classification (OA / mAcc)} &    \\
 \hline
 TempCNN$^*$ \cite{tempcnn} & 0.9 & 0.5 &  89.8 / 78.4 & 84.8 / 69.1 & 84.7 / 62.6  \\
 DuPLo$^*$ \cite{duplo} & 5.2 & 2.9 &  93.1 / 82.2  & 84.8 / 69.4 & 83.9 / 69.5 \\
 Transformer$^*$ \cite{transformer_sat} & 18.9 & 4.3 &  92.4 / 84.3  & 84.4 / 68.1  & 84.3 / 71.4  \\
 UNET3D \cite{Rustowicz2019SemanticSO}  & 6.2 & 11.2 &  92.7 / 83.9 & 84.8 / 70.2  &  84.8 / 71.4\\
 UNET2D-CLSTM \cite{Rustowicz2019SemanticSO}  & 2.3 & 35.5 & 93.0 / 84.0 &  84.7 / 70.3 & 84.7 / 71.6 \\
 U-TAE \cite{garnot_iccv} & 1.1 & 8.8 &  92.6 / 83.7  & 84.9 / 71.8 & 84.8 / 71.7 \\ 
 {\bf TSViT (ours)} & 1.7 & 11.8 &  {\bf 94.7 / 88.1}  & {\bf 87.1 / 75.5} & {\bf 87.8 / 74.2}  \\
\hline
\end{tabular}    
}
\caption{{\bf Comparison with state-of-the-art models from literature}. {\bf (top)} Semantic segmentation. {\bf (bottom)} Object classification. $^*$1D temporal only models. For each model we note its number of parameters ($\#$params. $\times 10^{6}$) and inference time (IT) for a single sample with T=52, H,W=24 and C=13 size input on a Nvidia Titan Xp gpu. We report overall accuracy (OA), mean intersection over union (mIoU) and mean accuracy (mAcc). For PASTIS we report results for fold-1 only; average test set performance across all five folds is shown in parenthesis for direct comparison with \cite{garnot_iccv}.}
\label{tab:results}
\end{table*}

\subsection{Ablation studies}\label{sec:ablation}
We perform an ablation study on design parameters of our framework using the Germany dataset \cite{Ru_wurm_2018}. Starting with a baseline TSViT with $L=4$ for both encoder networks, a single {\it cls} token, $h=w=3, t=1, d=128$ we successively update our design after each ablation. Here, we present the effect of the most important design choices; additional ablations are presented in the supplementary material.
% All results are presented in Table \ref{tab:ablation}. 
Overall, we find that the {\bf order of factorization} is the most important design choice in our proposed framework. Using a spatio-temporal factorization from the video recognition literature performs poorly at $48.8\%$ mIoU. Changing the factorization order to temporo-spatial raises performance by an absolute $+29.7\%$ to $78.5\%$ mIoU. 
% We note from Table \ref{tab:results} that without furtehr improthis model is enough to surpass the published sota in this dataset. 
Including {\bf additional \textit{cls} tokens} increases performance to $83.6\%$mIoU ($+5.1\%$), so we proceed with using $K$ {\it cls} tokens in our design. 
We test the effect of our date-specific {\bf position encodings} compared to a fixed set of values and find a significant $-2.8\%$ performance drop from using fixed size $\mathbf{P_T}$ compared to our proposed lookup encodings. 

Further analysis is provided in the supplementary material.
As discussed in section \ref{sec:tsvit_encoder} our spatial encoder blocks cross \textbf{\textit{cls}-token interactions}. Allowing interactions among all tokens comes at a significant increase in compute cost, $\mathcal{O}(K^2)$ to $\mathcal{O}(K)$, and is found to decrease performance by $-2.1\%$ mIoU.
Finally, we find that smaller {\bf patch sizes} generally work better, which is reasonable given that tokens retain a higher degree of spatial granularity and are used to predict smaller regions. Using $2\times2$ patches raises performance by $+1.2\%$mIoU to $84.8\%$ compared to $3 \times 3$ patches. 
Our final design which is used in the main experiments presented in Table \ref{tab:results} employs a temporo-spatial design with $K$ {\it cls} tokens, acquisition-time-specific position encodings, $2\times 2$ input patches and four layers for both encoders.

\vspace{-0.1cm}

\begin{figure}[!t]
\begin{center}
% \fbox{\rule{0pt}{1in} \rule{0.9\linewidth}{0pt}}
   \includegraphics[width=0.9\linewidth]{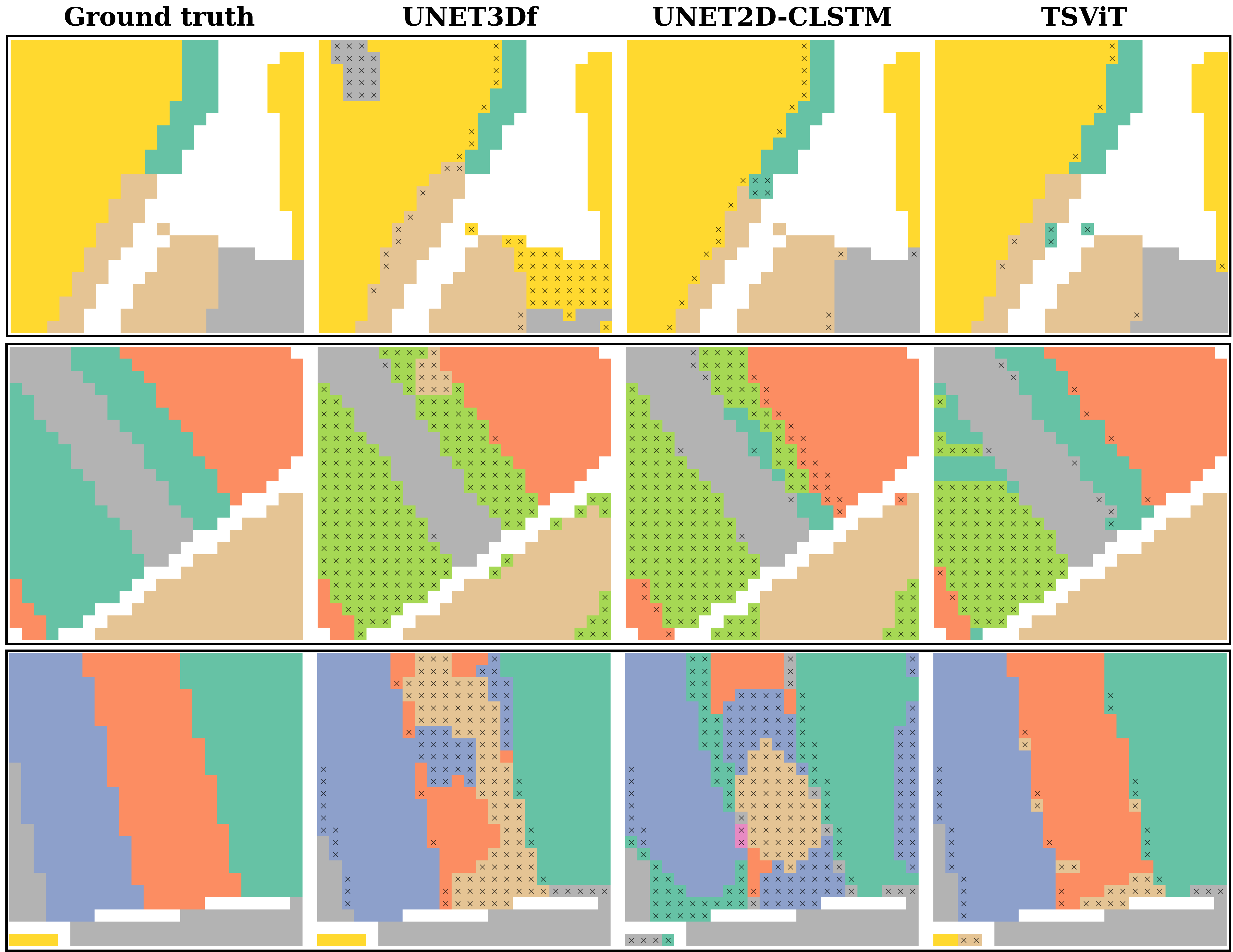}
\end{center}
   \caption{{\bf Visualization of predictions} in Germany. The background class is shown in black, "x" indicates a false prediction.}
\label{fig:sota_qualitative}
\end{figure}

% \vspace{-0.1cm}

\subsection{Comparison with SOTA}
In Table \ref{tab:results} and Fig.\ref{fig:overview}, we compare the performance of TSViT with state-of-the-art models presented in section \ref{sec:related_work}. For semantic segmentation, we find that all models from literature perform similarly, with the BiCGRU being overall the worst performer, matching CNN-based architectures only in T31TFM-1618. For all datasets, TSViT outperforms previously suggested approaches by a very large margin. A visualization of predictions in Germany for the top-3 performers is shown in Fig.\ref{fig:sota_qualitative}.
In object classification, we observe that 1D temporal models are generally outperformed by spatio-temporal models, with the exception of the Transformer \cite{transformer_sat}. Again, TSViT trained for classification consistently outperforms all other approaches by a large margin across all datasets. In both tasks, we find smaller improvements for the pixel-averaged compared to class-averaged metrics, which is reasonable given the large class imbalance that characterizes the datasets. 

\vspace{-0.30cm}

\section{Conclusion}\label{sec:conclusion}
In this paper we proposed TSViT, which is the first fully-attentional architecture for general SITS processing. Overall, TSViT has been shown to significantly outperform state-of-the-art models in three publicly available land cover recognition datasets, while being comparable to other models in terms of the number of parameters and inference time. However, our method is limited by its quadratic complexity with respect to the input size, which can lead to increased hardware requirements when working with larger inputs. While this may not pose a significant issue for semantic segmentation or SITS classification, it can present challenges for detection tasks that require isolating large objects, thus limiting its application. Future research is needed to address this limitation and enable TSViT to scale more effectively to larger inputs.

\noindent {\bf Acknowledgements} MT and EC acknowledge funding from the European Union's Climate-KIC ARISE grant. SZ was partially funded by the EPSRC Fellowship DEFORM: Large Scale Shape Analysis of Deformable Models of Humans (EP/S010203/1).

%%%%%%%%% REFERENCES
{\small
\bibliographystyle{ieee_fullname}
\bibliography{ms}
}

\end{document}